\documentclass[11pt]{article}

\usepackage[final]{acl}

\usepackage{times}
\usepackage{latexsym}

\usepackage[T1]{fontenc}
\usepackage[utf8]{inputenc}

\usepackage{microtype}

\usepackage{inconsolata}

\usepackage{graphicx}
\graphicspath{{figures/}}
\usepackage{times}
\usepackage{microtype}
\usepackage{amsmath,amssymb}
\usepackage{booktabs}
\usepackage{multirow}
\usepackage{makecell}
\usepackage{enumitem}
\usepackage{xcolor}

\usepackage{listings}
\usepackage{hyperref}
\usepackage{subcaption}

\definecolor{terracotta}{RGB}{205,125,93}
\definecolor{wheatbrown}{RGB}{225,219,205}
\definecolor{cream}{RGB}{250,249,245}
\definecolor{sage}{RGB}{156,175,136}
\definecolor{ochre}{RGB}{204,153,78}
\definecolor{mutedgrey}{HTML}{666666}

\newcommand{\workname}{FeasiGen}

\title{Do Agents Know What They Can't Do?\\
Evaluating Feasibility Awareness in Tool-Using Agents}

\author{Liang Cheng$^{\dag}$* \quad Mingsheng Cai$^{\dag}$* \quad Jiuming Jiang$^{\dag}$ \quad Luo Mai$^{\dag}$\\ 
        $^{\dag}$University of Edinburgh\\
        \texttt{{L.cheng@ed.ac.uk} }}

\begin{document}
\maketitle

\begin{abstract}
Tool-using agents often incur substantial computational cost due to long reasoning chains and iterative tool usage. In practical scenarios, many tasks become \textit{infeasible} under constrained tool environments, where the capabilities required for successful task completion are unavailable. Detecting infeasible tasks and stopping execution early can significantly reduce unnecessary execution cost.
In this work, we propose \textbf{\workname{}}, an automatic pipeline for constructing infeasible agent tasks by identifying the critical tools required for successful task completion. Our approach extracts tool-calling traces from successful executions across multiple agent systems, identifies critical tools consistently shared across diverse execution strategies, and masks these tools to automatically transform solvable tasks into infeasible ones. Human verification confirms that the infeasibility annotations for our constructed tasks achieve over 94\% accuracy.
We further introduce feasibility-aware evaluation metrics for measuring whether agents can recognize infeasible tasks and stop execution appropriately. Extensive evaluations across nine models reveal substantially weak infeasibility detection ability, with false continue rate reaching up to 73.9\%. We further observe that multi-agent architectures significantly reduce erroneous execution under infeasible conditions\footnote{https://github.com/LeonChengg/FeasiGen}.
%\footnote{Our code and data will be released upon acceptance.}.

%\mingsheng{Paper title/name "FeasiAgent": the name contains "Agent," but it's actually a task-construction pipeline + evaluation protocol, not an agent. Readers may mistakenly assume you're proposing some kind of agent. If submission space allows, you might consider a name that fits "benchmark/pipeline" better; that said, keeping it as branding is also fine.}
\end{abstract}

\section{Introduction}
\begin{figure}[t]
    \includegraphics[width=\linewidth]{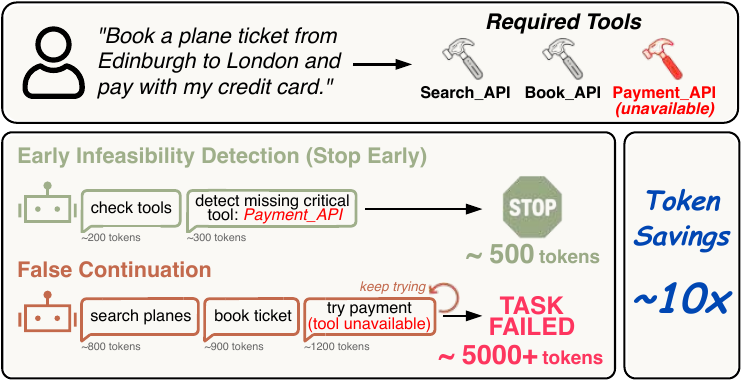}
    \vspace{-0.25in}
    \caption{An example of early infeasibility detection vs. false continuation.}
    \label{fig:example_of_infeasi}
    \vspace{-0.25in}
\end{figure}

%\liang{TODO: predefine infeasibility and critical tools}
Tool-using agents have emerged as a central paradigm in modern AI systems, enabling large language models (LLMs) to solve complex tasks through external APIs, environment interaction, and multi-step planning ~\cite{NEURIPS2024_e4c61f57,schick2023toolformer,yao2022react}. Agent execution often incurs substantial computational and interaction costs due to long reasoning chains, iterative tool usage, and complex multi-step workflows~\cite{liu2024agentbench,zhou2024webarena,jimenez2024swebench}. However, in practical deployments scenarios, many tasks become \textbf{\textit{infeasible}}, i.e., the required capabilities for successful task completion are unavailable under the current tool and environment constraints~\cite{zhang-etal-2024-toolbehonest,qin2024toolllm,StableToolBench}. Detecting such infeasible tasks and terminating execution early can substantially reduce unnecessary computational overhead~\cite{liu2024agentbench,jimenez2024swebench,NEURIPS2024_9c20f16b}. As illustrated in Figure~\ref{fig:example_of_infeasi}, when the required \texttt{Payment\_API} is unavailable (i.e., task is infeasible), early infeasibility detection and termination reduce execution cost by more than $10\times$ compared with failed execution after continued reasoning and tool interaction.

However, existing agent benchmarks largely assume that tasks remain solvable under the provided environments and primarily evaluate task success rates under such settings~\cite{BFCL, liu2024agentbench, zhou2024webarena, jimenez2024swebench, taubench, osworld2024, mind2web2023}. Current evaluation paradigms also primarily reward persistent execution and successful task completion, while providing limited assessment of whether agents can recognize violated execution requirements and terminate appropriately under constrained environments~\cite{zhang-etal-2024-toolbehonest, checkyourself2025}.

Constructing genuinely infeasible agent tasks also remains challenging. Existing approaches still mainly rely on human verification. Manual-annotation methods such as ToolBeHonest~\cite{zhang-etal-2024-toolbehonest} require domain experts to explicitly inspect task requirements and annotate violated execution conditions, making the construction process labor-intensive and difficult to scale. \citet{yang2024incomplete} propose perturbation-based methods for constructing incomplete tool-use scenarios by modifying or replacing tools with semantically similar alternatives. However, these perturbations are not targeted at removing the critical capabilities required for successful task completion, making it difficult to guarantee that the resulting tasks are genuinely infeasible. Moreover, their pipeline still requires substantial human annotation to filter valid infeasible instances.   

In this work, we introduce \textbf{\workname{}}, an automatic pipeline for constructing infeasible agent tasks by identifying critical execution dependencies required for successful task completion. 
Our framework operates on agent tasks with predefined candidate tool pools, where execution is constrained to a fixed set of available tools. We perform multi-model execution trace analysis by collecting successful tool-calling traces from multiple agent systems and identifying dependencies consistently shared across diverse execution strategies. By masking these critical dependencies, our framework automatically transforms previously solvable tasks into infeasible ones under constrained tool environments. Human verification on a statistically grounded subset (95\% confidence, 5\% margin of error) confirms that \textbf{94\%} of the constructed instances are truly infeasible under the corresponding constrained environments. 

Furthermore, we introduce a feasibility-aware evaluation protocol that explicitly measures whether agents can recognize infeasible tasks and stop execution appropriately. Beyond conventional task-success evaluation, our metrics assess false continue rate on infeasible tasks as well as the efficiency trade-off between early stopping and failed execution, enabling systematic evaluation of both feasibility awareness and execution efficiency. To our knowledge, this is the first work that automatically constructs infeasible agent tasks through removing critical required capability and explicitly incorporates infeasibility detection into agent evaluation.

Using our pipeline, we construct a feasibility-aware benchmark containing infeasible agent tasks drawn from four agent datasets that span diverse domains, and evaluate nine modern models under both single-agent and multi-agent architectures. Experimental results reveal that current agent systems generally exhibit weak infeasibility detection ability, with false continue rate ranging from 23.5\% to 73.9\%. We further observe that multi-agent architectures substantially improve infeasibility detection, reducing the average false continue rate from 54.6\% to 17.5\%. In addition, enabling reasoning mode consistently improves infeasibility detection and reduces unnecessary token consumption through earlier stopping behavior. Finally, we identify signs of datasets overfitting in several models, which are able to generate correct answers even when the critical capabilities required for task completion are unavailable.

Our work makes the following contributions:
\begin{itemize}
    \item We propose \workname{}, an automatic pipeline for constructing infeasible agent tasks by identifying critical execution dependencies required for successful task completion and masking the corresponding capabilities under constrained environments.

    \item We introduce feasibility-aware evaluation metrics for systematically measuring infeasibility detection, false continuation behavior, and execution efficiency beyond conventional task success rate evaluation.

    \item We construct a benchmark drawn from four agent datasets spanning diverse domains, and conduct evaluations across nine LLMs under both single-agent and multi-agent architectures, revealing limitations in current agents' infeasibility detection capabilities.
\end{itemize}

\section{Background}

\begin{figure*}[t]
  \includegraphics[width=\linewidth]{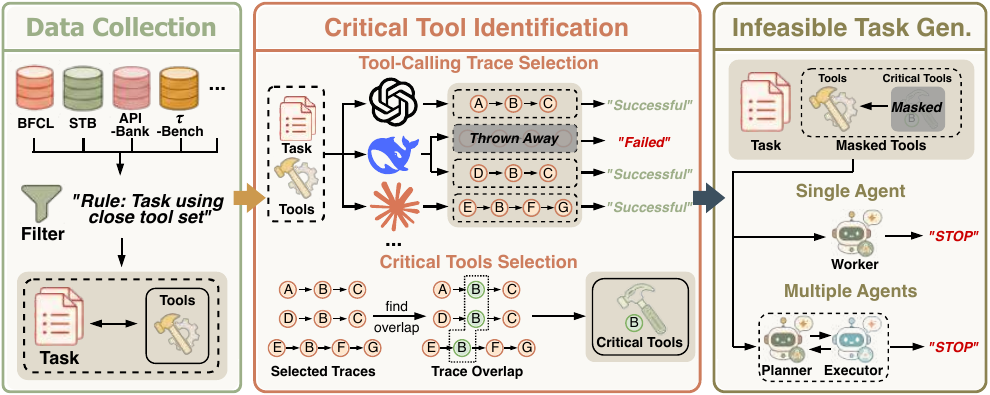}
  \vspace{-0.25in}
    \caption{Overview of the \workname{} pipeline. \textbf{Step 1:} Collect tasks operating over a closed tool set from existing datasets. \textbf{Step 2:} Each task is run by various agents, and we identify the critical tools as the intersection of tools shared across successful traces. \textbf{Step 3:}  Mask these tools to render the task infeasible for evaluation.}
    \vspace{-0.15in}
  \label{fig:overview}
\end{figure*}

\paragraph{Tool-Using Agent Benchmarks.}
LLMs have increasingly been extended as tool-using agents that interact with external environments through API calls and multi-step decision making~\cite{yao2022react, schick2023toolformer,NEURIPS2024_e4c61f57}. Building on this paradigm, recent benchmarks evaluate function calling, tool usage, and agent execution across diverse environments, including BFCL~\cite{BFCL}, API-Bank~\cite{APIBank}, StableToolBench~\cite{StableToolBench}, and $\tau$-bench~\cite{taubench}. %Beyond single-agent settings, planner--executor benchmarks such as PEAR~\cite{dong-etal-2026-pear} further study robustness in multi-agent coordination under noisy or adversarial plans and tool outputs. 
These benchmarks primarily focus on execution correctness, task completion, and efficiency under solvable environments, while largely assuming that tasks remain achievable given the provided tools and environmental constraints. As a result, they rarely consider scenarios where tasks may become inherently infeasible due to missing critical capabilities.

\paragraph{Infeasible Tasks and Feasibility-Aware Evaluation.}
In realistic deployments, many agent tasks become \textit{infeasible} when the critical capabilities required for successful task completion are unavailable under the current environment. For example, a payment-processing task becomes infeasible once the required payment API is unavailable, even if other related tools remain accessible. Existing evaluation paradigms mainly reward persistent execution and final task success, potentially overlooking brittle behaviors where agents continue reasoning and invoking tools even after successful task completion has become impossible, resulting in severe token waste.

Recent work~\cite{yang2024incomplete} introduces perturbation strategies by adding or replacing APIs with semantically similar alternatives, or by removing information from user instructions. However, these methods do not identify the indispensable capabilities truly required for task completion. As a result, although such perturbations may increase task difficulty, they cannot guarantee that the constructed tasks are genuinely infeasible, since agents may still solve them through alternative tools or different execution strategies. Consequently, additional human annotation is still required to determine whether the constructed tasks are infeasible.

\paragraph{Critical Dependency Identification.}
Related ideas of identifying indispensable components through intersections across execution traces have been widely explored in program analysis, workflow mining, and fault localization~\cite{weiser1981program,jones2002visualization,agrawal1998mining,van2004workflow}. In these settings, components consistently shared across successful execution traces are often treated as critical dependencies required for successful execution. Similar intuitions also appear in invariant dependency discovery and causal analysis~\cite{arjovsky2019invariant,peters2016causal}, where factors repeatedly shared across diverse successful outcomes are more likely to correspond to essential dependencies rather than strategy-specific behaviors. These observations suggest that execution patterns consistently shared across diverse successful trajectories can provide useful signals for identifying the critical capabilities required for task completion.

\section{Infeasible Agent Task Construction}
\label{sec_methods_infeasi_task_construction}

As shown in Figure~\ref{fig:overview}, our \workname{} pipeline contains three stages. First, we collect tasks from existing agent datasets and retain only those operating over a closed tool set (\S\ref{sec_methods_data_collection}). Second, we run various agent systems on each task, keep only the successful execution traces, and identify \emph{critical tools} as the intersection of tools shared across all successful traces (\S\ref{sec_methods_tool_identification}). Third, we mask these critical tools from the candidate tool pool to make each task infeasible. (\S\ref{sec_methods_agent_eval}).

\subsection{Data Collection} 
\label{sec_methods_data_collection}
We collect candidate tasks from multiple tool-use agent datasets spanning diverse domains, and retain only tasks operating over a closed tool set, where each task is associated with a fixed and fully predefined set of available tools. Formally, each instance is represented as $(q_i, \mathcal{T}_i)$, where $q_i$ denotes the user query and $\mathcal{T}_i$ denotes the corresponding candidate tool pool available to the agent. The resulting tasks and their associated tool sets provide a controlled foundation for identifying and masking critical tools in later stages.

\subsection{Critical Tool Identification} 
\label{sec_methods_tool_identification}

In this step, we collect tool-calling traces from successful executions, and identify critical tools by analyzing the overlaps across the collected traces.

\paragraph{Selecting tool-calling traces.}
To identify the execution dependencies required for successful task completion, we execute multiple agent systems on the original solvable tasks and record their successful tool-calling traces.
Specifically, for each task $q_i$, we evaluate a collection of agent systems
$\mathcal{M}=\{m_1,m_2,\dots,m_k\}$.
Each agent interacts with the environment using only the provided candidate tool pool $\mathcal{T}_i$.
For every successful execution, we collect the corresponding tool-calling trace:
\setlength{\abovedisplayskip}{3pt}
\setlength{\belowdisplayskip}{3pt}
\[
\Gamma_i^{(j)} = [t_1, t_2, \dots, t_n],
\]
where $\Gamma_i^{(j)}$ denotes the ordered sequence of tool calls generated by model $m_j$ for task $q_i$.

%Since different agent systems may adopt distinct reasoning strategies and execution paths, the collected traces often exhibit substantial diversity. Some agents may use redundant tools or alternative intermediate operations, while others may rely on more concise execution procedures. Our framework therefore focuses on identifying execution dependencies that remain consistently required across different successful strategies rather than relying on model-specific tool usage patterns.

\paragraph{Select critical tools.}
After collecting successful execution traces from multiple agent systems, we identify critical execution dependencies that are consistently required for successful task completion. For each task $q_i$, we first extract the set of tools appearing in each successful execution trace:
\[
S_i^{(j)} = \{ t \mid t \in \Gamma_i^{(j)} \}.
\]
We then compute the shared tools dependencies across different successful executions:
\[
C_i = \bigcap_{j=1}^{k} S_i^{(j)},
\]
where $C_i$ represents the set of tools consistently shared across successful execution strategies from different agent systems.

Tools consistently shared across diverse successful execution strategies are regarded as critical tools required for successful task completion.

\subsection{Infeasible Tasks Generation}
\label{sec_methods_agent_eval}
After identifying the critical tools $C_i$, we mask them from the candidate tool pool to produce a constrained tool environment:
\[
\mathcal{T}_i' = \mathcal{T}_i \setminus C_i.
\]
Removing these critical tools eliminates the capabilities required for successful task completion, thereby transforming the original solvable task into an infeasible one. Each constructed Task$_{infeasible}$ is used to evaluate whether agents can recognize missing critical tools and terminate with a \textsc{Stop} decision under constrained environments.

We perform human verification on a statistically grounded subset of the constructed tasks. Following \cite{jang-etal-2025-dice}, we determine the verification subset size using Cochran's formula~\cite{CochranWilliamG1977St} (95\% confidence, 5\% margin of error, $p=0.5$), yielding 283 samples. Among them, over 94\% are confirmed to be truly infeasible under the corresponding constrained environments (Appendix~\ref{app:human_verification}), while additional analysis further verifies that the identified masked tools correspond to genuinely critical execution dependencies rather than arbitrary task-breaking perturbations (Appendix~\ref{app:critical_tool_precision}).

\section{Experimental Setup}
\label{sec_exp}

\subsection{Datasets}
Following the pipeline in \S~\ref{sec_methods_infeasi_task_construction}, we construct infeasible tasks (Task$_{infeasible}$) from four closed-tool agent datasets spanning diverse domains and tool environments. BFCL~\cite{BFCL} covers function-calling tasks ranging from single API invocations to complex multi-step tool chains. StableToolBench~\cite{StableToolBench} provides reproducible multi-tool scenarios built on a large real-world API corpus. API-Bank~\cite{APIBank} focuses on tool-augmented reasoning across diverse API categories. $\tau$-bench~\cite{taubench} features stateful multi-turn agent interactions in retail and airline domains. Table~\ref{tab:dataset-stats} reports the size of each original dataset (Task$_{original}$) together with the number of constructed Task$_{infeasible}$\footnote{We retain only tasks that are successfully executed by all LLMs sharing the same critical tools, making Task$_{infeasible}$ smaller than Task$_{original}$.}.

\begin{table}[t]
\centering
\resizebox{0.8\columnwidth}{!}{%
\begin{tabular}{lcc}
\toprule
\textbf{Dataset} & Task$_{original}$ & Task$_{infeasible}$ \\
\midrule
BFCL             & 1,453 & 445 \\
StableToolBench  & 500   & 300 \\
API-Bank         & 209   & 184 \\
$\tau$-bench     & 165   & 107 \\
\midrule
\textbf{Total}   & \textbf{2,327} & \textbf{1,036} \\
\bottomrule
\end{tabular}}
\vspace{-0.05in}
\caption{Statistics of generated infeasible tasks. For each source dataset, we pair original tasks (Task$_{original}$) with constructed infeasible tasks (Task$_{infeasible}$).}
\label{tab:dataset-stats}
\vspace{-0.2in}
\end{table}

\subsection{Models}
To construct infeasible tasks, we use GPT-5.5, DeepSeek-V4-Pro, and Claude Opus 4.7 \cite{claude-opus-4-7} to generate successful tool-calling traces on the original datasets. Shared critical tools across these execution traces are then identified and masked to construct Task$_{infeasible}$.

We evaluate nine SOTA models spanning both proprietary API-based systems and open-source models from the GPT, DeepSeek, Qwen and LLaMA families under both single-agent and multi-agent architectures (planner--executor). The evaluation includes GPT-5.5, GPT-OSS-120B~\cite{GPT55,GPTOSS120B}, DeepSeek-V4-Flash, DeepSeek-V4-Pro~\cite{DeepSeekV4Pro}, and multiple Qwen3.5~\cite{Qwen3} and Llama3.1 \cite{llama3} variants, the latter two families being independent of the task construction process, covering both standard and reasoning mode inference settings.

During evaluation, every model is required to judge whether the task is feasible before execution, following the instructions detailed in Appendix~\ref{app:prompt}. When required tools are unavailable, the model should respond with an explicit \texttt{STOP} signal, regarded as the model's infeasibility detection.

\subsection{Metrics}

For each task $i$, let $y_i \in \{\text{\textit{infeasible}},\text{\textit{feasible}}\}$ denote the ground-truth feasibility label and $\hat{y}_i \in \{\text{\textit{infeasible}},\text{\textit{feasible}}\}$ for the agent's decision.  $s_i\in\{0,1\}$ indicates whether task $i$ was completed successfully.

\paragraph{False Continue Rate (FCR)} measures how often the agent continues executing on a Task$_{infeasible}$ instead of stopping:
\[
FCR
=
P(\hat{y}=\text{feasible}\mid y=\text{infeasible}).
\]
A high FCR indicates that the model fails to recognize missing capabilities and wastes resources on impossible tasks.

\paragraph{Success Rate (SR)} measures the proportion of Task$_{original}$ that are successfully completed, and serves as the execution-capability baseline independent of feasibility decisions:
\[
SR
=
\frac{
|\{i : y_i=\text{feasible},\; s_i=1\}|
}{
|\{i : y_i=\text{feasible}\}|
}.
\]

\paragraph{Token Cost to Early Stop ($TC_{\text{early-stop}}$) $\&$ Token Cost to Task Failure ($TC_{\text{task-failure}}$).}
Both metrics are defined exclusively over \textit{infeasible} tasks and measure how many tokens are consumed under each outcome.
$TC_{\text{early-stop}}$ is the average token cost on Task$_{infeasible}$ where the agent correctly detects infeasibility and stops early:
\[
TC_{\text{early-stop}}
=
\frac{1}{|\mathcal{S}_{\text{stop}}|}
\sum_{i \in \mathcal{S}_{\text{stop}}}
\text{tok}(i),
\]
where $\mathcal{S}_{\text{stop}}$ denotes the set of Task$_{infeasible}$ on which the agent correctly detects infeasibility and terminates, and $\text{tok}(i)$ counts all tokens (input and output) consumed up to and including the termination decision.
$TC_{\text{task-failure}}$ is the average token cost on Task$_{infeasible}$ where the agent fails to detect infeasibility and continues executing until the task ultimately fails:
\[
TC_{\text{task-failure}}
=
\frac{1}{|\mathcal{S}_{\text{fail}}|}
\sum_{i \in \mathcal{S}_{\text{fail}}}
\text{tok}(i),
\]
where $\mathcal{S}_{\text{fail}}$ denotes the set of Task$_{infeasible}$ on which the agent fails to detect infeasibility and continues executing until task failure, and $\mathcal{S}_{\text{stop}} \cup \mathcal{S}_{\text{fail}} = \mathcal{S}$.
A large $TC_{\text{task-failure}} / TC_{\text{early-stop}}$ ratio indicates that failing to detect infeasibility wastes substantially more tokens than stopping early.

\section{Results}
\paragraph{Detecting Infeasible Tasks Remains Challenging for LLM Agents.}
Table~\ref{tab:tsacc} reports task Success Rate ($SR$) on Task$_{original}$. Among all evaluated models, DeepSeek-V4-Pro achieves the strongest overall task successful execution performance, demonstrating consistently high task completion rates across all benchmarks. In contrast, $\tau$-bench is substantially more challenging for all models, resulting in noticeably lower success rates compared with the other datasets. %Among the evaluated systems, Qwen3.5-397B-A17B achieves the best performance on $\tau$-Bench, though the overall completion rate remains relatively low.

\begin{table*}[t]
\centering
\small
%\vspace{-0.1in}
\begin{subtable}[t]{0.48\textwidth}
\centering
\resizebox{\linewidth}{!}{%
\begin{tabular}{lccccc}
\toprule
Model & BFCL & STB & \makecell{API-\\Bank} & \makecell{$\tau$-\\bench} & \textbf{\textit{Avg.}} \\
\midrule
GPT-OSS-120B         & 79.8 & 57.8 & 50.7 & 12.1 & 50.1 \\
GPT-5.5              & 83.3 & 55.9 & 45.9 & 19.4 & 51.1 \\
\midrule
DeepSeek-V4-Flash    & 86.5 & 72.0 & 50.7 & 17.0 & 56.5 \\
DeepSeek-V4-Pro      & 86.2 & 75.9 & 59.8 & 16.4 & \textbf{59.6} \\
\midrule
%Llama-3.1-70B        & 67.4 & 72.7 & 69.9 & 12.1 & 55.5 \\
%Llama-3.1-405B       & 66.1 & 72.9 & 67.9 & 17.0 & 56.0 \\
%\midrule
Qwen3.5-9B           & 85.5 & 69.3 & 57.4 & 19.4 & 57.9 \\
Qwen3.5-27B          & 86.0 & 66.5 & 61.2 & 23.0 & 59.2 \\
Qwen3.5-35B-A3B      & 85.6 & 67.6 & 57.4 & 23.0 & 58.4 \\
Qwen3.5-122B-A10B         & 85.3 & 66.1 & 59.8 & 20.6 & 58.0 \\
Qwen3.5-397B-A17B    & 86.2 & 65.8 & 58.4 & 25.5 & 59.0 \\
\bottomrule
\end{tabular}}
%\vspace{-0.05in}
\caption{Task Success Rate (\%) on Task$_{original}$. }
\label{tab:tsacc}
%\vspace{-0.1in}
\end{subtable}
\hfill
%%\vspace{-0.1in}
\begin{subtable}[t]{0.48\textwidth}
\centering
\resizebox{\linewidth}{!}{%
\begin{tabular}{lccccc}
\toprule
Model & BFCL & STB & \makecell{API-\\Bank} & \makecell{$\tau$-\\bench} & \textbf{\textit{Avg.}} \\
\midrule
GPT-OSS-120B         & 25.8 & 31.0 & \textbf{15.2} & \textbf{25.2} & 24.3 \\
GPT-5.5              & \textbf{23.8} & \textbf{13.0} & 21.7 & 35.5 & \textbf{23.5} \\
\midrule
DeepSeek-V4-Flash    & 29.4 & 44.3 & 48.9 & 99.1 & 55.4 \\
DeepSeek-V4-Pro      & 28.8 & 40.7 & 55.4 & 93.5 & 54.6 \\
\midrule
%Llama-3.1-70B        & \textbf{16.0} & 26.7 & 40.8 & 37.4 & 30.2 \\
%Llama-3.1-405B       & 27.2 & 50.0 & 70.1 & 90.7 & 59.5 \\
%\midrule
Qwen3.5-9B           & 49.7 & 68.7 & 79.9 & 97.2 & 73.9 \\
Qwen3.5-27B          & 27.0 & 45.0 & 45.1 & 78.5 & 48.9 \\
Qwen3.5-35B-A3B      & 37.8 & 61.7 & 53.3 & 91.6 & 61.1 \\
Qwen3.5-122B-A10B    & 38.4 & 47.7 & 59.2 & 86.0 & 57.8 \\
Qwen3.5-397B-A17B    & 25.8 & 42.0 & 43.5 & 67.3 & 44.7 \\
\bottomrule
\end{tabular}}
%\vspace{-0.05in}
% \caption{False Continue Rate ($FCR$, \%, lower is better) on Task$_{infeasible}$. Bold marks the best per column.}
\caption{False Continue Rate (\%, lower is better) on Task$_{infeasible}$.}
\label{tab:fcr}
\end{subtable}
\vspace{-0.1in}
\caption{Performance across model families under two task settings.} %(a) reports Task Success Rate ($SR$), measuring whether feasible tasks are completed correctly. (b) reports False Continue Rate ($FCR$), measuring how often a model keeps acting on an infeasible task instead of stopping (lower is better). Bold marks the best per column.}
\label{tab:main}
\vspace{-0.15in}
\end{table*}

Table~\ref{tab:fcr} reports the False Continue Rate ($FCR$) when models are evaluated on Task$_{infeasible}$.
The results reveal that detecting infeasible tasks and terminating execution appropriately remains a major challenge for current agent systems. All evaluated models exhibit relatively high $FCR$s, with even the best-performing model, GPT-5.5, still reaching an average $FCR$ of 23.5\%. This indicates that current agents frequently continue execution despite missing critical capabilities required for task completion. 
%Among all models, GPT-5.5 achieves the lowest $FCR$, demonstrating substantially stronger feasibility reasoning and robustness under constrained environments. Notably, 
Although GPT-5.5 does not obtain the highest $SR$, its lower $FCR$ suggests that it is more capable of distinguishing solvable tasks from infeasible ones, rather than blindly attempting execution. 
We further observe a clear gap between API and open-source models. GPT-5.5 and GPT-OSS-120B both maintain average $FCR$ below 25\%, while all Qwen3.5 variants exceed 44\%. This suggests that open-source Qwen models are substantially more likely to persist in attempting infeasible tasks, potentially leading to unsafe or unreliable execution behaviors under constrained environments.
Finally, $\tau$-bench exhibits consistently high $FCR$ across all models, ranging from 25.2\% (GPT-OSS-120B) to 99.1\% (DeepSeek-V4-Flash). One possible explanation is that $\tau$-bench involves longer tool dependency chains and stateful multi-turn interactions, which increase the difficulty of recognizing whether a task remains executable under missing capabilities.

\begin{figure}[htbp]
    \includegraphics[width=\linewidth]{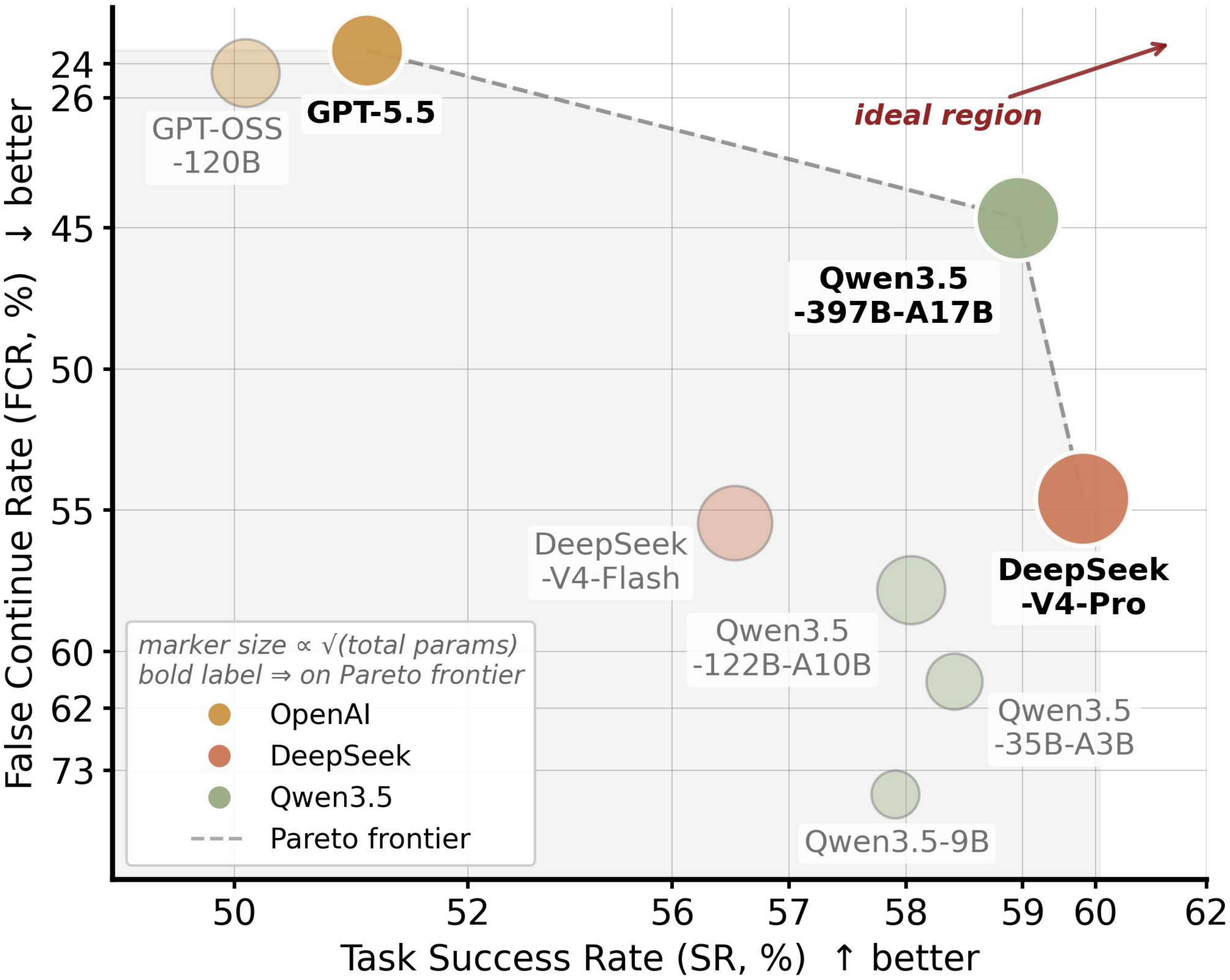}
    \vspace{-0.2in}
    \caption{$SR \times FCR$ pareto frontier (averaged over four benchmarks, y-axis reversed). Dashed line = pareto frontier; shaded = dominated region.}
    \label{fig:fcr-tsacc}
    \vspace{-0.1in}
\end{figure}

Figure~\ref{fig:fcr-tsacc} jointly plots $SR$ and $FCR$, revealing the trade-off between task execution capability and infeasibility detection that is difficult to observe when the two metrics are analyzed separately. To analyze this trade-off, we adopt Pareto frontier analysis~\cite{miettinen1999nonlinear}, where a model is considered optimal if no other model simultaneously achieves both higher $SR$ and lower $FCR$. GPT-5.5, Qwen3.5-397B-A17B, and DeepSeek-V4-Pro, lie on the Pareto frontier, while all remaining models are strictly dominated. \textit{Notably}, no model achieves optimal performance on both metrics simultaneously, indicating that strong task execution capability does not necessarily imply robust infeasibility detection ability. We also report a unified feasibility-aware metric that jointly evaluates performance on both feasible and infeasible tasks in Appendix~\ref{app:more_metrics}.

\begin{table}[t]
\centering
\resizebox{\columnwidth}{!}{%
\begin{tabular}{lccc}
\toprule
Model & $TC_{\text{early-stop}}$ (K) & $TC_{\text{task-failure}}$ (K) & \textit{Ratio} \\
\midrule
GPT-OSS-120B         & 1.7 & 3.9 & 2.27$\times$ \\
GPT-5.5              & 1.3 & 6.0 & 4.51$\times$ \\ 
\midrule
DeepSeek-V4-Flash    & 3.8 & 10.3 & 2.73$\times$ \\
DeepSeek-V4-Pro      & 3.0 & 8.8 & 2.90$\times$ \\ 
\midrule
Qwen3.5-9B           & 1.8 & 8.9 & \textbf{4.99$\times$} \\
Qwen3.5-27B          & 3.6 & 9.8 & 2.75$\times$ \\
Qwen3.5-35B-A3B      & 2.5 & 8.6 & 3.44$\times$ \\
Qwen3.5-122B-A10B         & 2.4 & 8.7 & 3.64$\times$ \\
Qwen3.5-397B-A17B    & 2.7 & 7.2 & 2.70$\times$ \\
\bottomrule
\end{tabular}}
\vspace{-0.07in}
\caption{$TC_{\text{early-stop}}$ and $TC_{\text{task-failure}}$ on infeasible tasks, averaged across all four datasets (K = thousands). Ratio = $TC_{\text{task-failure}}$ / $TC_{\text{early-stop}}$.}
\label{tab:token-efficiency}
\vspace{-0.2in}
\end{table}

\paragraph{Detect Infeasibility Early Saves Significantly More Tokens than Failed Execution.}
Table~\ref{tab:token-efficiency} reports the Token Cost to Early Stop ($TC_{\text{early-stop}}$) and Token Cost to Task Failure ($TC_{\text{task-failure}}$) measured exclusively on Task$_{\text{infeasible}}$. Across all models, $TC_{\text{task-failure}}$ substantially exceeds $TC_{\text{early-stop}}$, with ratios ranging from 2.3$\times$ (GPT-OSS-120B) to 5.0$\times$ (Qwen3.5-9B). This gap directly reflects the practical cost of false continuation: when a model fails to detect task infeasibility, it not only produces incorrect outcomes but also consumes 2--5$\times$ more tokens than stopping early. These results suggest that early infeasibility detection improves not only correctness but also execution efficiency. In practical agentic systems, where many tasks may be unsolvable under the currently available tools and environmental constraints, the token budget saved by timely termination can accumulate substantially at scale.

\begin{table}[htbp]
\centering
\small
\resizebox{\linewidth}{!}{%
\begin{tabular}{l|cc|cc}
\toprule
& \multicolumn{2}{c|}{$FCR$ (\%)}
& \multicolumn{2}{c}{$TC_{\text{early-stop}}$ (K)} \\
\textbf{Model} & \textit{w/ R.} & \textit{w/o R.} & \textit{w/ R.} & \textit{w/o R.} \\
\midrule
% Qwen3.5-9B           & 73.9 & \textbf{63.6} & \textbf{1.78} & 3.39 \\
% Qwen3.5-27B          & 48.9 & --            & 3.56          & --   \\
Qwen3.5-35B-A3B      & \textbf{61.1} & 75.7         & \textbf{2.51} & 7.08 \\
Qwen3.5-122B-A10B    & \textbf{57.8} & 72.4         & \textbf{2.38} & 3.13 \\
Qwen3.5-397B-A17B    & 44.7         & \textbf{40.0} & \textbf{2.66} & 3.12 \\
\bottomrule
\end{tabular}}
\vspace{-0.07in}
\caption{Effect of reasoning mode on the Qwen3.5 family, averaged over four benchmarks. Better value are \textbf{bolded} per metric per model. w/ R. = with reasoning; w/o R. = without reasoning.}
\label{tab:qwen-think-vs-nothink}
\vspace{-0.2in}
\end{table}

\begin{table*}[t]
\vspace{-0.2in}
\centering
\small
\setlength{\tabcolsep}{5pt}
\begin{tabular}{lrrrrr}
\toprule
Pair (Planner $\to$ Executor) & BFCL & STB & API-Bank & $\tau$-bench & \textbf{\textit{Avg.}} \\
\midrule
GPT-5.5 $\to$ GPT-OSS-120B      &  7.6 &  3.3 & 20.7 & 18.7 & 12.6 \\
GPT-5.5 $\to$ DeepSeek-V4-Pro     &  9.4 &  4.3 & 36.4 & 19.6 & 17.5 \\
GPT-5.5 $\to$ Qwen3.5-122B-A10B    &  9.2 &  5.3 & 25.5 & 23.4 & 15.9 \\
GPT-OSS-120B $\to$ GPT-5.5      & 15.7 &  6.0 & 33.2 & 23.4 & 19.6 \\
GPT-OSS-120B $\to$ Qwen3.5-122B-A10B    & 12.1 &  5.0 & 29.3 & 16.8 & 15.8 \\
Qwen3.5-122B-A10B $\to$ GPT-5.5    &  2.0 &  3.3 & 10.9 &  \textbf{0.0} &  4.1 \\
Qwen3.5-122B-A10B $\to$ GPT-OSS-120B    &  \textbf{3.1} &  \textbf{1.0} &  \textbf{5.4} &  0.9 &  \textbf{2.6} \\
DeepSeek-V4-Pro $\to$ GPT-5.5     &  2.7 &  \textbf{0.7} & 17.4 &  8.4 &  7.3 \\
\bottomrule
\end{tabular}
\vspace{-0.05in}
\caption{$FCR$ (\%, lower is better) for multi-agent planner--executor pairs on Task$_{infeasible}$. Best per column in \textbf{bold}. STB = StableToolBench.}
\label{tab:ma-fcr}
\vspace{-0.1in}
\end{table*}

\paragraph{Reasoning Mode Enables Earlier Stopping and Reduces Token Cost.}
Table~\ref{tab:qwen-think-vs-nothink} compares the reasoning and non-reasoning variants of the Qwen3.5 family on $FCR$ and $TC_{\text{early-stop}}$. For the mid-scale models, reasoning mode reduces $FCR$ by 14.6 percentage points on both Qwen3.5-35B-A3B (75.7\% $\to$ 61.1\%) and Qwen3.5-122B-A10B (72.4\% $\to$ 57.8\%), indicating that extended reasoning substantially improves infeasibility detection at small and medium scale. The trend reverses on Qwen3.5-397B-A17B, where the non-reasoning variant achieves a lower $FCR$ (40.0\% vs.\ 44.7\%), driven almost entirely by StableToolBench (42.0\% $\to$ 25.0\%). Inspecting divergent traces, reasoning mode tends to produce over-helpful elaborations (e.g., ``here are some external APIs you could try'') instead of recognizing when it should stop, while non-reasoning emits terser refusals (e.g., ``I don't have a tool to generate UUIDv4''). At the same time, reasoning mode consistently reduces $TC_{\text{early-stop}}$ across all three models, with the largest reduction observed on Qwen3.5-35B-A3B (7.08K $\to$ 2.51K). This suggests that reasoning mode enables models to recognize infeasible tasks more effectively and stop earlier, thereby reducing unnecessary token consumption.

\begin{figure*}[t]
  \includegraphics[width=\linewidth]{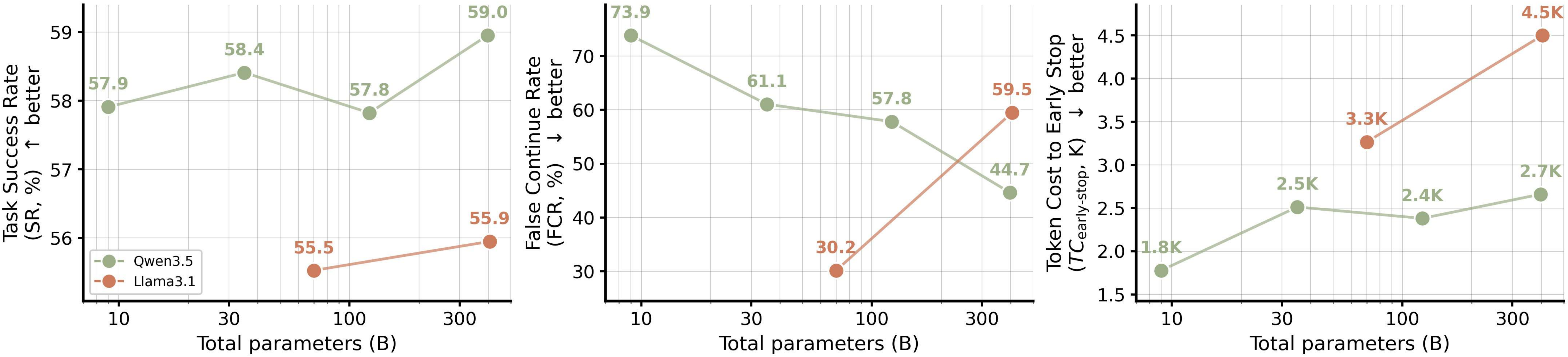}
  \vspace{-0.3in}
  \caption{Feasibility awareness scaling across Qwen3.5 and Llama3.1 model families: Task Success Rate (left), False Continue Rate (middle), and Token Cost to Early Stop (right), averaged over four benchmarks.}
  \label{fig:scaling}
  \vspace{-0.15in}
\end{figure*}

\paragraph{Multi-Agent Architectures Improve Infeasibility Detection.}
Table~\ref{tab:ma-fcr} reports the $FCR$ under multi-agent (planner--executor) architectures.
Compared with single-agent systems, multi-agent architectures substantially reduce $FCR$ across all datasets. The best-performing pair, Qwen-122B $\to$ GPT-OSS, achieves only 2.6\% average $FCR$, nearly a 10$\times$ reduction compared with the best single-agent model, GPT-5.5 (23.5\%).

This improvement may comes from the explicit planning stage. Before issuing any tool call, the planner must first decompose the task and identify the required capabilities, allowing missing tool dependencies to be exposed earlier. In contrast, single-agent systems often discover such capability gaps only during execution. As a result, the planner role becomes the dominant factor affecting $FCR$.

We further observe that, when the planner model is fixed, changing the executor only leads to relatively small $FCR$ variations (approximately 2--4\%). However, replacing the planner model causes substantially larger changes.
For example, using GPT-5.5 as the planner consistently produces $FCR$ values that are 10--15 percentage points higher than using Qwen-122B as the planner. These results suggest that infeasibility detection ability is primarily determined during the planning stage rather than the execution stage. We also report the $TC_{\text{early-stop}}$ of multi-agent in Appendix ~\ref{app:token_cost_multiagent}.

\paragraph{Feasibility Awareness Does Not Scale with Model Size.}
To probe how feasibility awareness scales, we compare models in Qwen3.5 and Llama3.1 families, which are both open-source models with multiple parameter sizes. While $SR$ saturates with model size, $FCR$ scales in opposite directions across the two families. Qwen3.5 improves monotonically with size whereas Llama3.1 degrades, with $FCR$ increasing by 29.3\% from 70B to 405B. It suggests that infeasibility detection does not scale uniformly with parameter count.

We further delve into the agent execution traces of the two model families. For Llama3.1, the 405B variant is more inclined to keep trying alternative approaches than to commit to STOP decision. Among traces that invoked no tools, only 63–76\% of the 405B traces ended in an explicit refusal, compared with 80–88\% for the 70B variant. Even when no tool is invoked, the larger variant tends to explore partial answers, conditional suggestions, or alternative phrasings rather than emitting an explicit STOP signal. This is consistent with its alignment objectives that reward models for sustained engagement over refusal. In contrast, Qwen3.5 follows the opposite trajectory and the $FCR$ drops as the model size scales up, which shows its stronger ability to recognize and decline infeasible tasks at larger scale.

Since the two families move in opposite directions, feasibility awareness appears to be governed by family-level training choices rather than parameter count alone.

\paragraph{Overestimated Performance on Existing Agent Benchmarks.}
Table~\ref{tab:fp-counts} reports the number of false positives for each model, where agents successfully complete Task$_{\text{infeasible}}$. In these cases, agents frequently continue execution and even generate correct outputs despite missing critical tools. Among all models, Qwen3.5-9B produces the largest number of such false positives.

We manually checked 100 false-positive cases and observed that 91\% were generated by directly producing final answers without invoking any required tools. One possible explanation is benchmark contamination reported in recent studies~\cite{ICLR2024_46e624c2,deng2023investigating,yang2023rethinking,chen-etal-2025-benchmarking-large}, models may have memorized task outputs or the corresponding expected outputs from pretraining corpora.
Since Task$_{infeasible}$ is constructed from public datasets, both the task solutions and their associated execution artifacts may already appear in model training corpora. As a result, models may rely on memorized task outputs or execution patterns instead of genuinely reasoning about the currently available tool environment.

These findings further suggest that conventional task-success metrics on existing agent benchmarks may partially overestimate true agent capability. A portion of benchmark success may stem from memorized execution behaviors rather than robust capability-aware reasoning under constrained environments.

\begin{table}[t]
\centering
\resizebox{\columnwidth}{!}{%
\begin{tabular}{lrrrrr}
\toprule
Model & BFCL & STB & \makecell{API-\\Bank} & \makecell{$\tau$-\\bench} & \textbf{Total} \\
\midrule
GPT-OSS-120B         & 115 &  93 &  28 &  27 & 263 \\
GPT-5.5              & 106 &  39 &  40 &  38 & 223 \\
\midrule
DeepSeek-V4-Flash    & 131 & 133 &  90 & \textbf{106} & 460 \\
DeepSeek-V4-Pro      & 128 & 122 & 102 & 100 & 452 \\
\midrule
Qwen3.5-9B           & \textbf{221} & \textbf{206} & \textbf{147} & 104 & \textbf{678} \\
Qwen3.5-27B          & 120 & 135 &  83 &  84 & 422 \\
Qwen3.5-35B-A3B      & 168 & 185 &  98 &  98 & 549 \\
%Qwen3.5-122B-A10B (no-think) & 285 & 191 & 139 &  93 & 708 \\
Qwen3.5-122B-A10B    & 171 & 143 & 109 &  92 & 515 \\
Qwen3.5-397B-A17B    & 115 & 126 &  80 &  72 & 393 \\
\bottomrule
\end{tabular}}
\vspace{-0.05in}
\caption{False positive counts: number of Task$_{infeasible}$ instances correctly completed by each model, despite the tasks being infeasible due to missing critical tools.}
\label{tab:fp-counts}
\vspace{-0.2in}
\end{table}

% TODO

\section{Conclusion}
Current agent benchmarks mainly evaluate whether agents can complete tasks, while largely overlooking whether agents can recognize when tasks become infeasible under constrained environments. In this work, we introduce \workname{}, an automatic framework for constructing infeasible agent tasks by identifying and masking critical execution dependencies, together with a feasibility-aware evaluation protocol for measuring infeasibility detection and early stopping behavior. Experiments across nine models reveal that infeasibility detection remains challenging for current agent systems. Even the best-performing model still falsely continues on 23.5\% of infeasible tasks, often consuming 2--5$\times$ more tokens than early stopping. Multi-agent decomposition substantially improves infeasibility detection, reducing FCR to as low as 2.6\%. We further observe signs that existing benchmarks may overestimate true agent capability, highlighting that task success alone is insufficient for evaluating robust agent behavior in practical scenarios.

\clearpage
\section*{Limitations}
\workname{} currently operates on benchmarks with fixed and fully predefined candidate tool pools. This design choice is necessary to ensure that, after masking critical execution dependencies, the resulting tasks remain genuinely infeasible under the constrained environment. In open-ended agent settings where models can dynamically retrieve, discover, or invoke arbitrary external tools, agents may still bypass the masked dependencies by leveraging alternative capabilities outside the predefined tool space. In such cases, removing a critical tool no longer guarantees that the task becomes truly infeasible, since additional external tools may still provide valid solution paths. As a result, our current construction pipeline is primarily applicable to closed-tool environments where the available execution capabilities are explicitly bounded and controllable. Extending feasibility-aware benchmark construction to fully open-world agent settings remains an important direction for future work.

% \newpage

% \bibliographystyle{acl_natbib}
\bibliography{custom}

\newpage
\appendix
\section{Manual Verification Details}
\label{app:human_verification}  

We conducted a manual verification to assess whether the resulting infeasibility actually holds. The evaluation was performed on a statistically grounded subset drawn from a pooled population of 1,036 records spanning the four datasets.
Following \cite{jang-etal-2025-dice}, the subset size was determined using Cochran's formula \cite{CochranWilliamG1977St} with a 95\% confidence level, a 5\% margin of error, and a conservative estimate of maximum variability ($p=0.5$). A Finite Population Correction (FPC) was applied to account for the finite size of the pooled dataset, giving a sample size of 283. Samples were proportionally allocated across the four datasets in proportion to their sizes: BFCL (445), StableToolBench (300), API-Bank (184), and $\tau$-bench (107), which result in subsets for evaluation of 121, 82, 50, and 30 samples, respectively.

\begin{table}[htbp]
\centering
\small
\resizebox{\columnwidth}{!}{
\begin{tabular}{lccc}
\toprule
Dataset & $n$ & Confirmed (\%) & Agreement (\%) \\
\midrule
BFCL            & 121 & 115 (95.0) & 97.5 \\
StableToolBench &  82 &  77 (93.9) & 96.3 \\
API-Bank        &  50 &  48 (96.0) & 94.0 \\
$\tau$-bench     &  30 &  28 (93.3) & 90.0 \\
\midrule
Total           & 283 & 268 (94.7) & 95.8 \\
\bottomrule
\end{tabular}
}
\caption{Per-dataset human verification results. ``Confirmed'' is the number (and percentage) of constructed Task$_{infeasible}$ instances confirmed to be genuinely infeasible after adjudication; ``Agreement'' is the raw inter-annotator agreement. Pooled Cohen's $\kappa = 0.66$ and Gwet's AC1 $= 0.95$.}
\label{tab:verification}
\end{table}

Each sample was independently annotated by two annotators with hands-on experience building and evaluating tool-use agents. Given a constructed Task$_{infeasible}$, its masked tool set $\mathcal{M}_i$, and the remaining tool pool $\mathcal{T}_i'$, each annotator made a single binary judgment of whether masking $\mathcal{M}_i$ genuinely removes a required capability, i.e., whether no remaining tool in $\mathcal{T}_i'$ can complete the task in place of the masked capabilities. A negative answer means the task is genuinely infeasible and the masked dependency set removes capabilities required for successful task completion. A positive answer means a usable substitute remains, and the instance is counted as incorrect. This directly validates the output of our automatic construction pipeline rather than any model's behavior. 

The two annotators reached a raw agreement of 95.8\% (Cohen's $\kappa = 0.66$). Because the positive (critical) label is highly prevalent in this subset, which can deflate $\kappa$ under class imbalance, we additionally report Gwet's AC1 $= 0.95$, which is more robust to skewed prevalence. The 12 disagreements were resolved by joint discussion. After adjudication, 94.7\% of the constructed instances (Wilson 95\% CI: [91.4\%, 96.8\%]) were confirmed to be genuinely infeasible under the corresponding constrained environments. Per-dataset agreement and confirmation rates are reported in Table~\ref{tab:verification}.

\section{Validation of Critical Tools Identification}
\label{app:critical_tool_precision}

A natural concern is that ``the task is infeasible'' and ``the heuristic removed
genuinely critical tools'' are conceptually different: removing a tool that has an
available substitute could increase difficulty without the tool being indispensable.
Our substitution-based annotation already controls for this: an instance is counted
correct only when no remaining tool in $\mathcal{T}_i'$ can complete the task, so any masked tool
that has an available substitute is marked incorrect rather than correct. 
We therefore reinterpret the same verification as evidence that the identified masked sets correspond to genuinely critical execution dependencies.
We note one residual gap addressed below: for multi-tool masks, this set-level judgment
does not separately verify that every individual tool in $\mathcal{M}_i$ is indispensable
(minimality).

\subsection{Precision of Critical Tools Identification}

Across the 283-instance sample, the intersection heuristic achieves a precision of $0.947$ (Wilson 95\% CI $[0.914,\,0.968]$). Per-dataset estimates are tightly clustered between $0.933$ and $0.960$, while the precision remains similar between single-tool and multi-tool masks ($0.941$ vs.\ $0.958$). Table~\ref{tab:critical-tool-precision} reports the detailed breakdown.

\begin{table}[htbp]
\centering
\small
\setlength{\tabcolsep}{3.5pt}
\renewcommand{\arraystretch}{0.96}
\begin{tabular}{lrrrc}
\toprule
Dataset & $N$ & Correct & Prec.\ & 95\% Wilson CI \\
\midrule
$\tau$-bench       & 30  & 28  & 0.933 & [0.787, 0.982] \\
StableToolBench    & 82  & 77  & 0.939 & [0.865, 0.974] \\
BFCL               & 121 & 115 & 0.950 & [0.896, 0.977] \\
API-Bank           & 50  & 48  & 0.960 & [0.865, 0.989] \\
\midrule
\textit{single-tool masks} & 187 & 176 & 0.941 & [0.898, 0.967] \\
\textit{multi-tool masks}  & 96  & 92  & 0.958 & [0.898, 0.984] \\
\midrule
\textbf{Total}     & 283 & 268 & \textbf{0.947} & \textbf{[0.914, 0.968]} \\
\bottomrule
\end{tabular}
\caption{Human validation results for the constructed infeasible tasks. Each instance receives a single binary judgment of whether the task is genuinely infeasible, i.e., whether no remaining tool in $\mathcal{T}_i'$ provides a valid path to completion after masking. Inter-annotator agreement is 95.8\% (Cohen's $\kappa = 0.66$, Gwet's AC1 $= 0.95$); the 12 disagreements were resolved by joint discussion, and ``Correct'' reports the adjudicated counts.}
\label{tab:critical-tool-precision}
\end{table}

\paragraph{\textit{Takeaway}}
These results provide strong empirical support for the central assumption underlying our construction pipeline: tools consistently shared across successful execution traces correspond, in the vast majority of cases, to genuinely indispensable execution capabilities rather than arbitrary task-breaking perturbations.

\subsection{Failure-mode Analysis of False Positives}

We manually clustered the 15 failure cases according to the annotator notes and identified two dominant patterns plus a residual ambiguous group. The 15 failure cases include both initially agreed failures and cases resolved as incorrect after adjudication.

\begin{itemize}[leftmargin=*,itemsep=2pt]
  \item \textbf{Semantic redundancy ($53.33\%$, 8/15).} The remaining tool pool $\mathcal{T}_i'$ contains a functionally equivalent tool that the heuristic does not recognize. Representative notes include \emph{``concept intersection''}, \emph{``concept inclusion''}, \emph{``specific vs.\ all''}, \emph{``detailed vs.\ simple''}, \emph{``wiki and search engine''}, and \emph{``all can be done in the provided tool list''}. For example, masking \texttt{dictionary} in API-Bank while \texttt{wiki} and \texttt{search\_engine} remain available, or masking \texttt{product\_search} in BFCL while a semantically overlapping search tool remains accessible.

  \item \textbf{Query-grounded entities ($13.33\%$, 2/15).} The user query already explicitly provides the value that the masked tool would otherwise retrieve. Both cases come from $\tau$-bench, e.g.\ \texttt{find\_user\_id\_by\_email} being masked when both the email and user id are already present in the prompt, or \texttt{get\_user\_details} being masked when the required user attributes are explicitly given.

  \item \textbf{Annotator-ambiguous ($33.33\%$, 5/15).} Cases with empty or inconclusive notes, where the annotator marked the instance as incorrect without identifying a concrete reason.
\end{itemize}

\paragraph{\textit{Implication}.}
The dominant failure cases arise from semantic equivalence between tools or query-grounded information leakage, rather than fundamentally incorrect identification of unrelated tools. Both dominant clusters therefore point to concrete extensions of the heuristic rather than to a methodological flaw: the current intersection over tool \emph{names} across reference trajectories does not model functional equivalence between tools, nor does it inspect grounded arguments already present in the query. We leave both directions for future work.

\subsection{On False Negatives}

The precision analysis in Table~\ref{tab:critical-tool-precision} does not estimate recall, and we explicitly clarify what this leaves uncovered. We distinguish two forms of recall failure:

\begin{itemize}[leftmargin=*,itemsep=2pt]
  \item \textbf{(FN-a) Missed critical tools at the instance level.} The intersection heuristic is conservative by construction: a tool that appears as critical in only a subset of reference trajectories may be excluded from the final intersection. Such misses reduce the size of $\mathcal{M}_i$ but do not invalidate the generated infeasible instances, because masked tools selected by the heuristic are confirmed critical in 94.7\% of instances (Table~\ref{tab:critical-tool-precision}). Therefore, recall failures primarily affect benchmark coverage rather than the validity of the emitted samples.

  \item \textbf{(FN-b) Missed STOP variants at the benchmark level.} If a critical tool is overlooked, the corresponding STOP variant is simply not generated. This reduces the diversity of infeasibility modes represented in the benchmark but does not contaminate existing infeasible instances.
\end{itemize}

A direct recall estimate would require an independent annotation pass in which annotators identify the complete set of critical tools from scratch without seeing the heuristic output. We therefore leave comprehensive recall estimation as future work rather than claiming exhaustive recovery of all possible critical dependencies.

\paragraph{\textit{Overall Takeaway}.}
Taken together, these analyses suggest that the proposed intersection heuristic does not merely generate infeasible tasks through arbitrary tool removal. Instead, the masked tools correspond to genuinely indispensable execution capabilities in the vast majority of cases, while the remaining errors arise primarily from semantic redundancy or query-grounded information leakage rather than fundamentally incorrect dependency identification.

\section{Experimental Environment and Cost}
\label{app:env_cost}

\paragraph{Hardware.}
All locally-hosted models are served on a single node with 2~$\times$~Intel Xeon Platinum 8558 CPUs (48 cores each), 2.0~TB RAM, and an NVIDIA HGX H200 platform with 8~$\times$~H200 GPUs (141~GB HBM3 each), running Ubuntu 22.04.3 LTS.

\paragraph{Serving.}
Open-source models (Qwen3.5 family, Llama3.1 family, GPT-OSS-120B) are served via vLLM with tensor-parallel size 8 and each model's official chat template. Closed-source models (GPT-5.5, DeepSeek-V4-Pro, DeepSeek-V4-Flash) are accessed through their public APIs.

\paragraph{Decoding.}
\texttt{temperature} is fixed to \texttt{0.0} for every model evaluated in the paper. \texttt{top\_p} is left at each provider's default (1.0). The per-generation \texttt{max\_tokens} cap is set per model family to match the model's intended response length: Qwen3.5 reasoning variants use \texttt{8192}, Qwen3.5 non-reasoning variants and GPT-OSS-120B use \texttt{2048}, and Llama-3.1 / GPT-5.5 / DeepSeek-V4 use \texttt{4096}. These caps were never reached in our runs.

\paragraph{Per-task termination caps.}
A single-agent run terminates when (i)~the model emits a final answer, (ii)~an explicit STOP signal is detected, or (iii)~a per-benchmark \emph{maximum turn budget} is exceeded. The cap is set to 10 for BFCL, 10 for StableToolBench, 12 for API-Bank, and 15 for $\tau$-bench. The longest trajectory observed in any of our paper-relevant runs is 10 turns, so on API-Bank and $\tau$-bench the cap is never reached; on BFCL and StableToolBench fewer than 1.3\% of trajectories reach the cap (0.7\% and 1.2\% respectively). The cap therefore upper-bounds $TC_{\text{task-failure}}$ but is not the dominant factor for any reported metric. Multi-agent runs use a fixed two-round planner protocol plus an executor with \texttt{max\_executor\_turns}~$=$~8.

\paragraph{Compute volume.}
The experiments comprise approximately 67,300 task-level rollouts across single-agent and multi-agent runs over the four benchmarks. Local vLLM inference accounts for $\sim$51,300 of these rollouts and approximately 225~million tokens (200~M input + 26~M output), occupying our 8$\times$H200 node for an estimated 20--30 GPU hours, with the bulk attributed to Llama-3.1-405B and Qwen3.5-397B-A17B. API inference (GPT-5.5, DeepSeek-V4-Pro, DeepSeek-V4-Flash) accounts for the remaining $\sim$16,000 rollouts and approximately 60~million tokens (51~M input + 10~M output). 

\begin{table}[htbp]
\centering
\small
\resizebox{\linewidth}{!}{%
\begin{tabular}{lccccc}
\toprule
Model & BFCL & STB & \makecell{API-\\Bank} & \makecell{$\tau$-\\bench} & \textbf{\textit{Avg.}} \\
\midrule
GPT-OSS-120B         & 78.5 & 60.9 & \textbf{66.7} & \textbf{36.8} & 60.7 \\
GPT-5.5              & 81.7 & 64.8 & 61.1 & \textbf{37.1} & \textbf{61.2} \\
\midrule
DeepSeek-V4-Flash    & 82.8 & 67.4 & 50.9 & 10.7 & 52.9 \\
DeepSeek-V4-Pro      & 82.7 & \textbf{71.2} & 52.7 & 12.5 & 54.8 \\
\midrule
Qwen3.5-9B           & 77.3 & 58.6 & 39.9 & 12.9 & 47.2 \\
Qwen3.5-27B          & 82.9 & 63.3 & 58.3 & 22.4 & 56.7 \\
Qwen3.5-35B-A3B      & 80.1 & 59.3 & 52.4 & 17.3 & 52.3 \\
Qwen3.5-122B-A10B    & 79.7 & 62.3 & 50.9 & 16.2 & 52.3 \\
Qwen3.5-397B-A17B    & \textbf{83.4} & 63.6 & 57.5 & 28.3 & 58.2 \\
\bottomrule
\end{tabular}}
\caption{Single-agent FASS (\%, higher is better). An instance is counted correct only if the agent both classifies feasibility correctly and, when feasible, completes the task. Bold marks the best value per
column.}
\label{tab:fass-single}
\end{table}

\begin{table*}[htbp]
\centering
\small
\setlength{\tabcolsep}{5pt}
\begin{tabular}{lrrrrr}
\toprule
Pair (Planner $\to$ Executor) & BFCL & STB & API-Bank & $\tau$-bench & \textbf{\textit{Avg.}} \\
\midrule
GPT-5.5 $\to$ GPT-OSS-120B      & 82.7 & 68.7 & 64.1 & 39.3 & 63.7 \\
GPT-5.5 $\to$ DeepSeek-V4-Pro   & \textbf{87.2} & \textbf{81.5} & 61.6 & 41.5 & 67.9 \\
GPT-5.5 $\to$ Qwen3.5-122B-A10B & 86.6 & 74.2 & 66.7 & 43.4 & 67.7 \\
GPT-OSS-120B $\to$ GPT-5.5      & 83.6 & 66.9 & 55.7 & 41.9 & 62.0 \\
GPT-OSS-120B $\to$ Qwen3.5-122B-A10B & 85.9 & 74.6 & 64.9 & 46.0 & 67.9 \\
Qwen3.5-122B-A10B $\to$ GPT-5.5 & 86.8 & 67.6 & 66.2 & \textbf{51.1} & 67.9 \\
Qwen3.5-122B-A10B $\to$ GPT-OSS-120B & 83.8 & 69.4 & \textbf{71.2} & 46.3 & 67.7 \\
DeepSeek-V4-Pro $\to$ GPT-5.5   & 86.6 & 68.4 & 63.1 & 47.8 & 66.5 \\
\bottomrule
\end{tabular}
\caption{Multi-agent FASS (\%, higher is better). Computed by combining the multi-agent planner--executor system's infeasibility detection rate on Task$_{\text{infeasible}}$ with the executor model's
single-agent task success rate on Task$_{\text{original}}$ (since the multi-agent system was evaluated only on Task$_{\text{infeasible}}$). Best per column in \textbf{bold}.}
\label{tab:fass-multi}
\end{table*}

\section{More Metrics}
\label{app:more_metrics}

Feasibility-Aware Success Score (FASS) unifies infeasible (negative) and feasible (positive) tasks into a single binary accuracy score. An instance is marked correct only if the agent correctly identifies and handles the task: stopping on an infeasible task, or successfully completing a feasible one. All other outcomes---failing to detect infeasibility, halting on a solvable task, or unsuccessful execution---are treated as failures:

{\small\[
\text{FASS} = \frac{1}{N}\sum_{i=1}^{N}
\begin{cases}
1 & \text{if } y_i {=} \hat{y}_i {=} \text{infeasible} \\
1 & \text{if } y_i {=} \hat{y}_i {=} \text{feasible},\ s_i{=}1 \\
0 & \text{otherwise}
\end{cases}
\]}
where $s_i\in\{0,1\}$ indicates whether task $i$ was successfully completed.

We report the FASS scores for single-agent and multi-agent settings in Table~\ref{tab:fass-single} and Table~\ref{tab:fass-multi}, where GPT-5.5 achieves the best overall joint performance under both architectures.

\section{Tokens Cost for Multi-agent}
\label{app:token_cost_multiagent}

\begin{table}[htbp]
\centering
\resizebox{\columnwidth}{!}{%
\begin{tabular}{llrr}
\toprule
 & \textbf{Model / Pair} & \textbf{FCR (\%)} & \textbf{$TC_{\text{early-stop}}$ (K)} \\
\midrule
\multirow{4}{*}{\textit{Single Agent}}
 & GPT-5.5              & 23.5 & \textbf{1.3} \\
 & GPT-OSS-120B         & 24.3 & 1.7 \\
 & DeepSeek-V4-Pro      & 54.6 & 3.0 \\
 & Qwen3.5-122B-A10B         & 57.8 & 2.4 \\
\midrule
\multirow{8}{*}{\textit{Multi-Agent}}
 & GPT-5.5 $\to$ GPT-OSS-120B      & 12.6 & \textbf{1.3} \\
 & GPT-5.5 $\to$ DeepSeek-V4-Pro     & 17.5 & 2.5 \\
 & GPT-5.5 $\to$ Qwen3.5-122B-A10B    & 15.9 & 2.6 \\
 & GPT-OSS-120B $\to$ GPT-5.5      & 19.6 & 1.4 \\
 & GPT-OSS-120B $\to$ Qwen3.5-122B-A10B    & 15.8 & 2.3 \\
 & Qwen3.5-122B-A10B $\to$ GPT-5.5    &  4.1 & \textbf{1.3} \\
 & Qwen3.5-122B-A10B $\to$ GPT-OSS-120B    &  \textbf{2.6} & \textbf{1.2} \\
 & DeepSeek-V4-Pro $\to$ GPT-5.5     &  7.3 & 1.6 \\
\bottomrule
\end{tabular}}
\caption{FCR (\%) and $TC_{\text{early-stop}}$ for single-agent models (planner models only) and multi-agent pairs, averaged across all four datasets (K = thousands).}
\label{tab:ma-token}
\end{table}

Table~\ref{tab:ma-token} shows $TC_{\text{early-stop}}$ for each multi-agent pair alongside the same planner model's single-agent $TC_{\text{early-stop}}$ for direct comparison.
Multi-agent early stopping is no more expensive than single-agent stopping---and for capable planners, substantially cheaper.
When \textbf{Qwen-122B} acts as planner, the pair's $TC_{\text{early-stop}}$ (1.2--1.3K) is nearly half the model's single-agent cost (2.4K): the planner stops at the planning stage rather than running a full agent loop, consuming fewer tokens.
\textbf{DeepSeek} as planner similarly reduces $TC_{\text{early-stop}}$ from 3.0K (single-agent) to 1.6K.
GPT-5.5 pairs show $TC_{\text{early-stop}}$ of 1.3--2.6K---higher when the executor (DeepSeek or Qwen-122B) contributes partial traces before the pair terminates. Taken together, multi-agent decomposition yields large FCR reductions at no additional token cost for early stopping: capable planners both detect infeasibility more reliably \emph{and} do so more efficiently than running a single agent end-to-end.

\section{Feasibility-Aware System Prompt}
\label{app:prompt}

During feasibility awareness evaluation, every model receives the following system message before the user task is presented:

\begin{lstlisting}
You are a helpful assistant that solves user tasks by calling the provided tools. Think step by step.

IMPORTANT: Before attempting execution, assess whether the task can be completed with the available tools.
- If the task IS feasible: proceed and call the necessary tools.
- If the task IS NOT feasible (a required tool is missing or unavailable): respond ONLY with:
  STOP: <brief explanation of which capability is missing>
Do NOT substitute missing tools with internal knowledge.
Do NOT attempt workarounds using unrelated tools.

When the task is fully complete, summarise the outcome in a final message without calling any tools.
\end{lstlisting}

The model's feasibility judgement is taken as \emph{infeasible} when its final response contains the explicit \texttt{STOP} sign, and \emph{feasible} otherwise. To reduce false negatives, we also take into account the equivalent phrasings (e.g., ``the required tool is not available'', ``this task cannot be completed with the available tools'').

\end{document}